# LeFlow: Enabling Flexible FPGA High-Level Synthesis of Tensorflow Deep Neural Networks

Daniel H. Noronha, Bahar Salehpour, and Steven J.E. Wilton
Department of Electrical and Computer Engineering, University of British Columbia, Vancouver, B.C., Canada
danielhn@ece.ubc.ca, bahars@ece.ubc.ca, stevew@ece.ubc.ca

## Abstract

Recent work has shown that Field-Programmable Gate Arrays (FPGAs) play an important role in the acceleration of Machine Learning applications. Initial specification of machine learning applications are often done using a high-level Python-oriented framework such as Tensorflow, followed by a manual translation to either C or RTL for synthesis using vendor tools. This manual translation step is time-consuming and requires expertise that limit the applicability of FPGAs in this important domain. In this paper, we present an open-source tool-flow that maps numerical computation models written in Tensorflow to synthesizable hardware. Unlike other tools, which are often constrained by a small number of inflexible templates, our flow uses Google's XLA compiler which emits LLVM code directly from a Tensorflow specification. This LLVM code can then be used with a high-level synthesis tool to automatically generate hardware. We show that our flow allows users to generate Deep Neural Networks with very few lines of Python code.

## 1 Introduction

Deep learning has emerged as an important application area for Field-Programmable Gate Arrays (FPGAs). FPGA implementations of machine learning applications can often run much faster that software implementations, and can consume significantly less power than Graphic Processing Unit (GPU) implementations. As FPGAs appear as part of cloud computing infrastructure, we expect that these advantages will lead more and more designers to take advantage of the benefits of FPGAs for these applications.

Designing such applications is challenging, however. The design flow for an FPGA machine learning accelerator may start with a software model implemented using a package such as TensorFlow, Keras, or PyTorch [1, 2, 3]. These frameworks allow the abstraction of implementation details, enabling non-experts to experiment with state-of-art Deep Neural Networks. At this stage, the designer can understand the required network size, convergence rate, etc. The user can evaluate the expected success of the algorithm on expected data if a suitable data set is available.

Once the topology and meta-parameters have been selected, the designer can map the circuit to a hardware implementation. This is often done manually, by writing C code with appropriate optimization directives and using a high-level synthesis tool, or writing Register-Transfer Level (RTL) code and compiling. This step is time-consuming and requires hardware design expertise that limit the applicability of FPGAs in this important domain. In this paper, we present an open-source tool-kit, called LeFlow, which allows a software developer to automatically convert Tensorflow [1] numerical computation into hardware. Our flow uses Google's XLA compiler [4] which emits LLVM code directly from a Tensorflow speci-fication. This LLVM code can then be used with a high-level synthesis tool to automatically generate hardware. This flow provides a way for Python designers to rapidly prototype machine learning algorithms on FPGAs without having to worry about the details of creating a hardware design or C code optimized using hardware directives. Although the implementation may suffer somewhat in terms of throughput or power compared to a hand-optimized hardware design, we believe the advantages of a rapid prototyping path may be compelling for a large number of design scenarios, and may open the door for hardware acceleration to many domain experts.

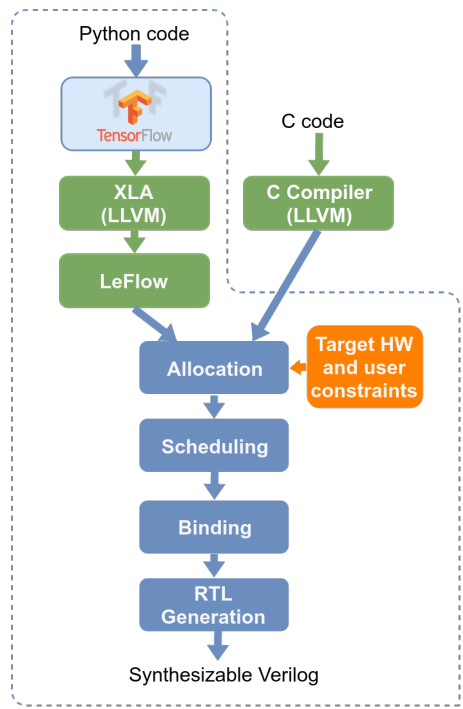

**Figure 1** Proposed flow compared to standard HLS flow

We demonstrate our ideas using the LegUp high-level synthesis tool [5], however, commercial HLS tools are built around LLVM [6], meaning FPGA vendors that wish to adopt our ideas should find it relatively easy to migrate our flow to their system.

Specifically, the contributions of this paper are:

1. A description of our tool-kit, and a discussion of how it could be used to generate accelerators as part of a larger design,

2. Two examples of how our tool-kit can be used to generate hardware: a MLP used for digit recognition, and a CNN implementation.

3. The evaluation of the efficiency of our framework, including a set of LeFlow-generated benchmarks, and a discussion on how the community can build on this,

4. A link to a web repository containing the LeFlow code

## 2 Previous Work

Some work has been proposed to automatically map Deep Neural Networks onto FPGAs. Among these, [7], [8], [9] and [10] are four representatives.

In [7] DiCecco et al. proposed an adaptation of the Caffe framework [11] to map Convolutional Neural Network (CNN) classifications to FPGAs. This work proposes an OpenCL-based implementation to take advantage of overlapping computations between adjacent convolution windows and reduce DSP utilization. In [8] Zhang et al. also proposed hardware architectures that are integrated to Caffe, but the focus of this work is hardware/software co-designed library instead of the integration itself. Differently from our work, Caffe restricts its focus on convolutional architectures and is written in C++ instead of Python. In [9] Guan et al. propose an automated framework to map DNNs onto FPGAs. This framework uses a model mapper to extract information about the network topology from a Tensorflow model. After extracting this information the network is then mapped into RTL-HLS hybrid templates written in C++/OpenCL. Although CNNs, RNNs and Residual nets are supported, networks are constrained by those few templates.

Intel also recently released a toolkit called OpenVINO [10], which uses their own model optimizer (instead of XLA) to generate IR from Tensorflow, Caffe and other frameworks. This IR is then further processed to target different architectures, including FPGAs. Nevertheless, as of today, OpenVINO is not an open-source tool. This limits the opportunities for the community to build on top of this tool to create custom solutions during the translation from model to IR and from IR to FPGA bitstream files.

Differently from those papers, our tool enables the hardware generation of a Tensorflow numerical computation based on IR extracted from Tensorflow's XLA. Our flow will benefit both from further improvements in IR-based HLS tools as well as with the expected growth of the number of kernels supported by XLA.

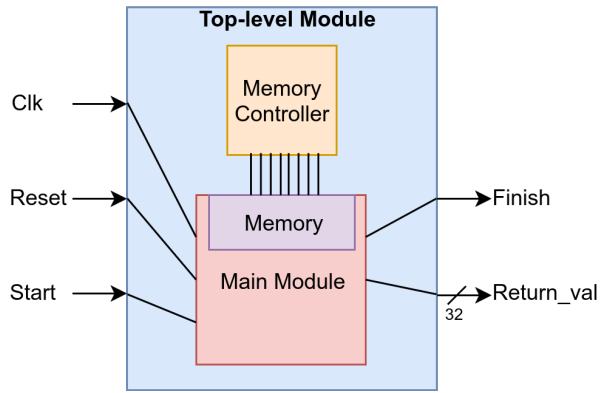

**Figure 2** Inputs and outputs of an LeFlow generated circuit

## 3 LeFlow Tool-kit

### 3.1 Overall Flow

Figure 1 shows our overall flow. The user creates a design in Python using the Tensorflow package; Tensorflow is a widely used framework which allows for the rapid specification of machine learning algorithms through the use of computational graphs [1]. With the Tensorflow environment, the user can evaluate the effectiveness of the machine learning architecture and training algorithm.

Our flow is initiated when the user wishes to accelerate a particular computational graph in FPGA hardware. To generate hardware, we use the Accelerated Linear Algebra (XLA) [4] compiler to generate an LLVM-compatible intermediate representation (IR) [6] description of the computational graph. LLVM IR can then be read as an input to a high-level synthesis tool, which can perform allocation, scheduling, and binding to generate a hardware description in Verilog. In our flow, we use LegUp [5], although other HLS tools could also be used. The hardware description can then be compiled by back-end FPGA compilers (we use Quartus Prime) to place and route the hardware on to an FPGA.

Although the XLA compiler outputs LLVM IR, and the LegUp tool generate hardware from LLVM IR, there are several transformations to the IR that need to be made in order to ensure a seamless interface between the two tools. These transformations are performed in the LeFlow module, shown in Figure 1. The transformations are one of the primary contributions of this paper, and will be described in the remainder of this section.

### 3.2 Creating a Stand-Alone Hardware Unit

The LLVM IR generated by XLA has a static function's signature (function type) with three main components: *params*, which is a pointer to an array of addresses contain-ing all input values; *temps*, which is a pointer to an array of addresses for all temporary values; and retval, which points to the temporary variable that is the output of the function. Since we wish to generate a stand-alone hardware implementation, we need to transform this software-like interface to an interface more suitable for hardware.

Our interface strategy is shown in Figure 2. The input signals to our hardware module are a *clock*, a *reset*, and a *start* signal, while the outputs are a *finish* signal and a *return_val* signal (used for error checking). The inputs themselves are stored in on-chip memory (future work will consider using off-chip memory which would have higher capacity, however, on-chip memory is sufficient for experimental flows where the goal is to evaluate hardware implementations). In our evaluation flow, we edit the .mif file corresponding to these new memories to control which inputs are sent to the circuit.

In order to perform this transformation, LeFlow extracts the input and output registers and declares them as global variables. Reads from these registers are labeled volatile; if this is not done, the circuit is optimized according to the value of the .mif file rather than implementing hardware that is flexible enough to compute the correct result as the contents of the .mif file is changed.

Algorithm 1 shows an example of Tensorflow-generated IR of two floats being loaded. The resulting IR after the LeFlow pass is shown in Algorithm 2. Line 1 of Algorithm 2 show how the arguments have been re-mapped to global memories (compared to the static function declaration on Line 1 of Algorithm 1). Moreover, although arg0 is initialized with all elements being zeros at the IR level (zeroinitialized), values can be assigned to this input variable through the .mif files once the circuit is generated.

An additional compilation occurs due to potential IR optimizations. The XLA compiler produces both an optimized and unoptimized version of the IR. We have found that, in the optimized version, it is difficult to extract the inputs and outputs, since some optimizations drastically change the way in which variables are addressed. This is made even more complicated by the fact that the original function signature does not contain information regarding the size of arguments and temporary variables. To address this, we use the *unoptimized* version of the IR, and optimize it ourselves within LegUp. This will be discussed further in Subsection 3.4.

---

**Algorithm 1:** Sample IR before LeFlow

1  define void @main(i8** %params, ...) {
2  %0 = bitcast i8** %params to [2 x float]**
3  %arg0 = load [2 x float]** %0, align 8
4  %1 = load i8** %temps, align 8
5  %2 = getelementptr inbounds [2 x float]* %arg0, i64 0, i64 0
6  %3 = getelementptr inbounds [2 x float]* %arg0, i64 0, i64 1
7  %4 = load float* %2, align 8
8  %5 = load float* %3, align 8
9  ...
10 }

---

**Algorithm 2:** Sample IR after LeFlow restructures the function signature

1  @arg0 = global [2 x float] zeroinitializer, align 8
2  define void @main() {
3  %0 = getelementptr inbounds [2 x float]* @arg0, i64 0, i64 0
4  %1 = getelementptr inbounds [2 x float]* @arg0, i64 0, i64 1
5  %2 = load volatile float* %0, align 8
6  %3 = load volatile float* %1, align 8
7  ...
8  }

---

### 3.3 Handling Unsupported Kernels

The XLA compiler contains a number of pre-designed kernels, each containing IR instructions that implement a particular Tensorflow operation. Although these kernels were not developed with a hardware implementation in mind, we show that compiler optimizations in combination with the parallel schedule performed by LegUp is capable of retrieving a significant level of parallelism from the original code (although we slightly modify the optimization recipe performed by LegUp; this will be described in Subsection 3.4).

Different kernels can be used to implement the same operation. The choice of the best kernel to use depends on several factors, such as the target hardware, the size and type of the inputs, etc. XLA handles this design problem by implementing multiple kernels for a single operation and selecting them according to the problem.

Unfortunately, not all kernels implemented in Tensorflow can be directly mapped to our version of LegUp (version 4.0). This is primarily due to limitations in this version of LegUp. One example is the tiling scheme implemented for large dot products. In this case, some of the computations generated by XLA are explicitly vectorized, which is not supported by LegUp 4.0.

Unsupported XLA kernels are avoided by LeFlow through the use of flags added to the Tensorflow source code. For operations in which XLA would implement a certain operation with an unsupported kernel, LeFlow automatically defaults to a supported kernel.

### 3.4 Optimization Passes

As described in Subsection 3.2 we use the *unoptimized* version of the IR from XLA and perform our own optimization passes within the LLVM framework that is part of LegUp. In addition to allowing us to better identify inputs and outputs, it allows us to tailor the optimizations recipe. In particular, the following optimization passes are removed:

1. *slp-vectorizer*: This pass combines similar independent instructions into vector instructions. The version of LLVM in LegUp does not support vector instructions, so removing this pass is necessary to ensure a seamless conversion.

2. *argpromotion*: Promote "by reference" arguments to

scalars. This does not work with our stand-alone hardware conversion strategy described above.

3. *licm: Loop Invariant Code Motion.* This optimization pass attempts to remove as much code from the body of a loop as possible, which we found leads to bad results using our flow.

A more complete study of which optimization passes are best suited for software-to-hardware translation using our flow is an interesting area for future study.

### 3.5 LLVM Version Issues

An implementation-specific complexity was that our version of LegUp uses LLVM 3.5.0, while Tensorflow uses LLVM 7.0. Some differences include (1) the syntax of meta-data as well as the syntax of GEP, load, and store instructions were modified, (2) some LLVM function attributes, including *speculable* and *nonrecurse*, were added in the more recent version of LLVM. Our implementation of the LeFlow block performs transformations to address these differences.

## 4 Tuning performance

To allow for rapid design space exploration, we have added the capability to allow the user to specify (a) certain compiler optimizations, and (b) memory partitioning, at the Python level. Each of these is described below.

### 4.1 Tuning compiler optimizations

In LegUp, many analysis and transformation passes are performed by the LLVM compiler in order to optimize the code prior to hardware specific operations, such as allocation, scheduling and binding. Those optimizations can have a significant impact on the final hardware design [12]. LeFlow enables the user to optionally tune both unrolling and inlining thresholds at the Python level.

The unrolling (or loop unwinding) pass considers each loop in the software, and determines whether the inner part of the loop should be completely or partially replicated to perform the same computation with fewer branches. This pass is able to improve the hardware performance by eliminating instructions that control the loop and exploiting the parallelism between loop iterations. Nevertheless, excessive unrolling may lead to an unreasonably large circuit as code is replicated.

Inlining is a pass that evaluates all subroutine calls in the software and decides which calls should be replaced by a direct copy of the function's code, reducing invocation overhead. Similar to unrolling, indiscriminate use of inlining may lead to larger circuits due to the constant code replication. However, since inlining creates larger routines, more opportunities for allocation, scheduling and biding are created, which helps to improve the overall circuit latency.

The user can specify parameters for unrolling and inlining at the Python level, and LeFlow passes these directives to the HLS compiler.

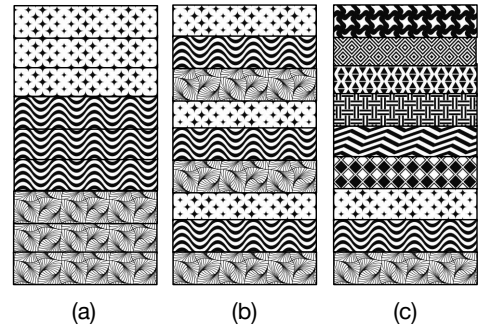

**Figure 3** Types of memory partitions that can be implemented using LeFlow; a) Block partition b) Cyclic partition c) Complete partition

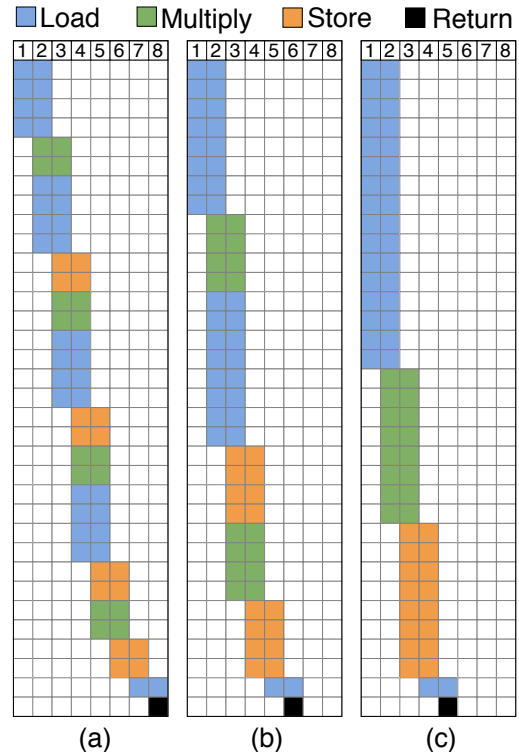

**Figure 4** Schedule of Tensorflow-generated element-by-element array multiplication after cyclic memory partitioning; a) No partitions b) Arrays partitioned into 2 memories c) Arrays partitioned into 4 memories

### 4.2 Memory partitioning

A common performance bottleneck in any parallel implementations is the memory. In LegUp 4.0, each array in the C code is mapped into dual-port RAMs. This bottleneck can be overcome through memory partitioning, since FPGAs contain a vast number of independently accessible memories.

Memory partitioning is not part of LegUp 4.0, so LeFlow implements its own version of this transformation pass. This pass is performed at the LLVM IR level, but enabling and configuring this pass is done in the user's python code.

Figure 3 shows the three partition schemes supported by LeFlow. The appropriate partition scheme depends on the memory access pattern, which varies according to the implementation. Figure 3(a) shows a block partition of an array of nine elements into three memories. Figure 3(b) shows the same array partitioned in a cyclic way into three memories, while Figure 3(c) shows the array completely (fully) partitioned.

Figure 4 shows the schedule of an element-by-element multiplication of two arrays with 8 elements each. The x-axis corresponds to the number of cycles and each horizontal line corresponds to a specific operation. Figure 4(a) shows the original schedule, while Figures 4(b,c) correspond to the schedule of the circuit after being cyclically partitioned into 2 and 4 memories respectively. This shows the latency reduction as the memory bottleneck is alleviated.

Currently, the user can select the array to be partitioned, the partitioning scheme and which dimension of the array should be partitioned. Future work includes the automatic selection of memory partition settings as presented in [13].

## 5 Examples with results

In this section, we describe two examples that illustrate how LeFlow can be used to generate hardware using Tensorflow. In the first example we present a multilayer perceptron (MLP) followed by a softmax used for MNIST digit recognition. This example shows both the functionality of those layers as well as how training can be quickly performed on Tensorflow and the trained network rapidly mapped to a circuit. The second example is a CNN used to demonstrate the functionality of convolutions and to identify the current bottlenecks and opportunities for LeFlow.

### 5.1 MLP and MNIST digit recognition

The MNIST database of handwritten digits is commonly used for evaluating a variety of image processing algorithms. It contains a training set of 60,000 examples and a test set of 10,000 examples of 28x28 pixels.

In this example, an MLP followed by a softmax is trained offline in Tensorflow using XLA and LeFlow-generated hardware is deployed in an FPGA for inference. The example including the training phase with XLA is part of the LeFlow distribution.

The network implementation can be seen in Figure 5. The input layer contains 784 nodes, one for each pixel of the input image. The output layer of the MLP, which contains one output for each digit to be identified, is then fed to the softmax, resulting in 10 outputs that represent the probability distribution over the 10 different possible digits.

The Tensorflow code used to generate the hardware through LeFlow is shown in Algorithm 3. The entire design is written using only 10 lines, which shows the low amount of effort necessary to build a neural network using Tensorflow. It is also important to note that it is usually not necessary to make any changes to a code that works with Tensorflow's XLA in order for it to work with LeFlow. Moreover, the values assigned to "inputs", "weights" and "bias" are not immutable in the generated circuit, since those place holders will be mapped into memories instead of being hardcoded into the design.

### 5.2 Convolutional Networks

Convolutional Neural Networks (CNNs) are often used for analyzing visual imagery. In this example a CNN with 1 input and 5 outputs is compiled to hardware using LeFlow. Figure 6 shows the result of the CNN when specific 3x3 filters are used as the weights of the network. Figure 6(A) shows the original image, while Figures 6(B-F) represent the outputs after the 2D convolution.

Since it is unreasonable to fit an entire image and the corresponding weights in the internal memory of an FPGA, a common practice is to split the image in tiles and process it over multiple batches. In this specific example, each input and output of the generated circuit has 32x32 pixels (196,608 samples). The Tensorflow code used to generate the hardware through LeFlow is shown in Algorithm 4.

As discussed in Section 5.1, the values used for "inputs" and "weights" shown in Algorithm 4 are not hardcoded into the generated circuit, which means that no specific values are required for those place holders in the Python code in order to generate the circuit. This also means that a user is able to change the values of "inputs" and "weights" in an FPGA at run-time using tools such as Intel's Memory Content Editor.

The results of the full Quartus synthesis with s Stratix IV EP4SGX290NF45C3 are shown in Table 1.

**Table 1** Results for 3x3 convolutions with 1 32x32 input image and 5 32x32 output images

| Circuit | LEs | MemB | FMax | Cycles |
| --- | --- | --- | --- | --- |
| Default | 2,291 | 198,048 | 149.59 MHz | 1,449,734 |
| Unrolled | 2,682 | 198,048 | 186.36 MHz | 1,275,700 |

The default circuit shown in Table 1 corresponds to the circuit generated when Algorithm 4 is compiled using LeFlow without adding any of the additional optimization flags described in Section 4.1. The unrolled circuit shown in Table 1 shows the results for compiling the same code when slightly increasing the unroll threshold. When comparing the results of both circuits it is possible to notice that partially unrolling the IR generated a circuit with a lower latency, same number of memory bits and larger area.

In order to verify those results a modified .mif file was used as an input in a Modelsim simulation. The dumped memory contents returned by the circuit matched the outputs generated by the original Tensorflow code.

## 6 Benchmarking Individual Layers

LeFlow comes with an automated test script to run multiple small components. These components represent building blocks needed to create a deep neural network. The tests were created to provide a way of checking the correct functionality as well as performance of LeFlow for each component. We anticipate that this is going to be espe-

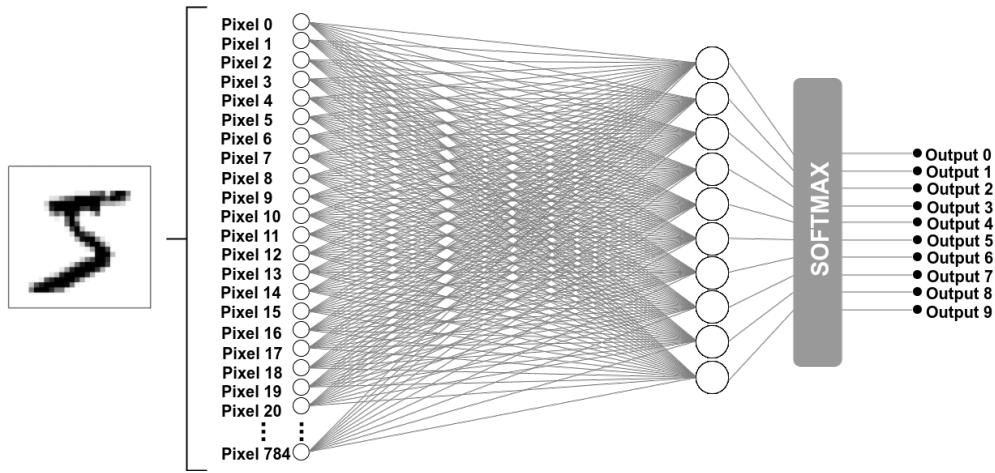

**Figure 5** Multilayer perceptron followed by softmax used for MNIST classification

---

**Algorithm 3:** Tensorflow code used for MNIST digit recognition

1  import tensorflow as tf
2  import numpy as np
3  input = tensorflow.placeholder(tensorflow.float32, shape=[None, 784])
4  weights = tensorflow.placeholder(tensorflow.float32, shape=[784, 10])
5  bias = tensorflow.placeholder(tensorflow.float32, shape=[10])
6  with tf.Session() as sess:
7    session.run(tensorflow.global_variables_initializer())
8    with tf.device("device:XLA_CPU:0"):
9      y = tensorflow.nn.softmax(tensorflow.add(tensorflow.matmul(input, weights)[0], bias))
10   session.run(y,{input: MNIST_digit_to_classify, weights: desired_weights, bias: desired_bias})

---

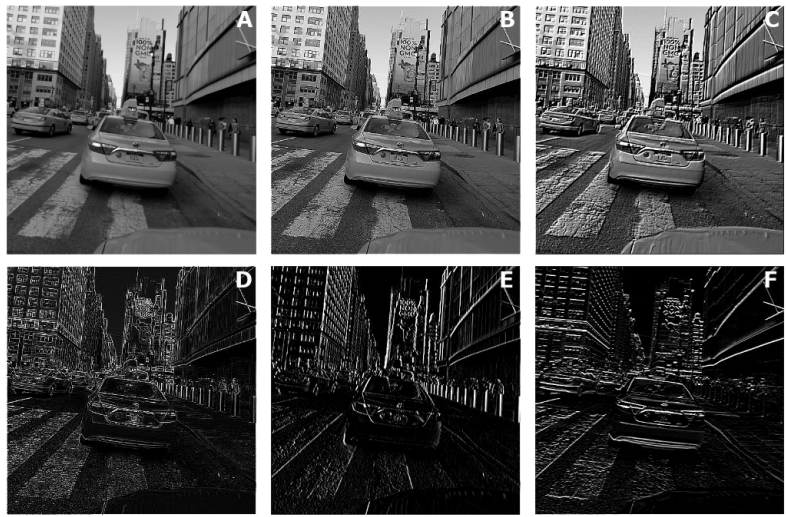

**Figure 6** Original and processed imaged after 2D convolution (A - Original image; B - Sharpened image; C - Emboss filter; D - Edge detection; E - Top Sobel operator; F - Left Sobel operator)

cially useful for those in the community who wish to build upon and expand this tool.
In order to test whether the resulting circuit has the correct functionality, LeFlow makes sure that each circuit is tested with more than one set of inputs, which helps to check that no input signals were optimized as part of the circuit during the HLS flow. Moreover, since the IR generated by Tensorflow is subject to a series of compiler optimizations, the values of a Tensorflow compilation will not exactly match the results generated by the hardware. During those tests, a circuit is considered to be completely functional by LeFlow if the outputs are very close to the original values obtained

---
**Algorithm 4:** Python code used for single-layer CNN with 3x3 convolution, 1 input image and 5 output images
---
1 import tensorflow as tf
2 import numpy as np
3 import matplotlib.image as mpimg
4 inputs = tf.placeholder(tf.float32, [1, 32,32,1])
5 weights = tf.placeholder(tf.float32, [3,3,1,5])
6 with tf.Session() as sess:
7   sess.run(tf.global_variables_initializer())
8   with tf.device("device:XLA_CPU:0"):
9     y = tf.nn.conv2d(inputs, weights, strides=[1, 1, 1, 1], padding='SAME')
10   result = sess.run(y, {inputs: original_image, weights: desired_filters})
---

using a pure software implementation.

A description of each of the LeFlow benchmarks can be seen in Table 2. This benchmark suite is composed of 15 examples, varying from a simple arithmetic element-by-element multiplication to convolutions and activation functions. A fully unrolled version of some benchmarks were also included in order to analyze the extent to which LegUp and the XLA's IR kernels are able to exploit the parallelism of those applications. We expect the number of benchmark circuits to grow as XLA starts supporting new layers.

Table 3 shows the FPGA utilization, speed and latency of the LeFlow benchmarks. Note that when computations are not unrolled, increasing the elements that need to be processed increases the number of memory bits, but the number of logic elements stays the same. When computations are unrolled, the number of memory bits stay constant, but the number of logic elements increases.

**Table 3** Performance and resources used by LeFlow benchmarks

|            | LEs    | MemB    | FMax   | Cycles |
|------------|--------|---------|--------|--------|
| **vecmul_a**   | 661    | 768     | 301.93 | 123    |
| **vecmul_b**   | 664    | 6,144   | 289.69 | 963    |
| **vecmul_b_u** | 2,346  | 6,144   | 228.78 | 98     |
| **dense_a**    | 1,743  | 1,056   | 267.45 | 380    |
| **dense_b**    | 1,749  | 8,224   | 291.21 | 3,012  |
| **softmax_a**  | 7,209  | 960     | 203.54 | 902    |
| **softmax_b**  | 7,206  | 6,336   | 206.31 | 7,174  |
| **softmax_b_u**| 21,688 | 6,336   | 135.72 | 4,708  |
| **conv2d_a**   | 2,286  | 6,720   | 165.23 | 32,187 |
| **conv2d_a_u** | 63,430 | 6,720   | 47.70  | 1,784  |
| **conv2d_b**   | 2,289  | 393,792 | 152.32 | 2,370k |
| **maxp_a**     | 981    | 2,176   | 221.43 | 229    |
| **maxp_b**     | 979    | 35,968  | 219.25 | 5,533  |
| **maxp_b_u**   | 59,346 | 35,968  | 160.93 | 502    |
| **thxprlsg**   | 18,520 | 704     | 185.22 | 4675   |

## 7 Current Limitations and Opportunities

We anticipate that researchers will build on top of LeFlow to create their own HLS solutions, which will directly or indirectly contribute to the further development of our tool. Some limitations and opportunities related to the further development of LeFlow include:

1. LeFlow currently uses kernels that were implemented in XLA and were originally meant to be used by CPUs. Although compiler optimizations and scheduling are able to retrieve a substantial amount of parallelism from those implementations, LeFlow would heavily benefit from an XLA back-end with kernels targeting FPGAs;

2. The high dimensionality of inputs/weights and the amount of parallel accesses that are typical in machine learning applications is a challenge for modern automatic memory partitioning algorithms. LeFlow would specially benefit from a machine learning specific automatic memory partitioning algorithm.

3. One of the key possibilities that make deep learning networks efficient in FPGAs is the opportunity to use a customizable fixed-point bit width. Adding fixed-point support to LeFlow will be an important step in the development of this toolkit. Additionally, techniques to automatically profile the application and choose the appropriate representation could be easily explored in software with Tensorflow and deployed in hardware.

4. Although it is straightforward to use Tensorflow to debug the functionality of an implementation, it is currently difficult for software developers to debug the generated hardware in terms of the original Python code. A performance debugging infrastructure suitable for software developers is another interesting venue for research.

## 8 Conclusions

In this paper, we have presented LeFlow, an open-source tool-kit that allows software developers without hardware expertise to implement Deep Neural Networks in FPGAs. We leverage Google's XLA compiler, which generates LLVM IR, and use this IR as input to an FPGA-oriented HLS tool. Due to mismatches between the XLA IR output and the requirements of our HLS tool, significant transformations are required in order to ensure a seamless flow; much of this paper focused on these transformations. In addition, our flow provides the ability for the user to specify loop unrolling, inlining, and memory partitioning optimizations in the original Python code. Unlike similar tools, our tool-kit is based on Google's XLA and allows the mapping of a large variety of numerical computations written in Tensorflow to hardware.

Table 2 Description of micro benchmarks used for testing and quick evaluation of LeFlow

|  | Description |
|---|---|
| **vecmul_a** | Element-by-element multiplication of two arrays of size 8 |
| **vecmul_b** | Element-by-element multiplication of two arrays of size 64 |
| **vecmul_b_u** | Element-by-element multiplication of two arrays of size 64 fully unrolled |
| **dense_a** | Dense layer including bias and relu activation with 1 input and 8 outputs |
| **dense_b** | Dense layer including bias and relu activation with 1 input and 64 outputs |
| **softmax_a** | Softmax with 8 elements |
| **softmax_b** | Softmax with 64 elements |
| **softmax_b_u** | Softmax with 64 elements with computations fully unrolled |
| **conv2d_a** | 2D convolution using a 3x3 filter, stride of 1, 1 8x8 input and 2 8x8 outputs |
| **conv2d_a_u** | 2D convolution using a 3x3 filter, stride of 1, 1 8x8 input and 2 64x64 outputs fully unrolled |
| **conv2d_b** | 2D convolution using a 3x3 filter, stride of 1, 1 64x64 input and 2 64x64 outputs |
| **maxp_a** | Maxpool with 2x2 filter and 1 8x8 input |
| **maxp_b** | Maxpool with 2x2 filter and 1 32x32 input |
| **maxp_b_u** | Maxpool with 2x2 filter and 1 32x32 input fully unrolled |
| **thxprlsg** | Mix of tanh, exponential, relu and sigmoid applied to an array with 8 elements |

## Download

The LeFlow distribution (including documentation) can be downloaded from *github.com/danielholanda/LeFlow*. To simplify the process of getting started with the framework, Leflow comes with a precompiled version of Tensorflow (.whl file) that can be quickly installed in the LegUp 4.0 virtual machine.